\documentclass[runningheads]{llncs}
\usepackage{xcolor}
\usepackage{attachfile2}
\usepackage{graphicx}
\usepackage{amssymb}
\usepackage{multirow}
\usepackage{rotating}
\usepackage{amsmath}
\usepackage{pdfpages}
\usepackage{svg}
\begin{document}

\title{SCOL: Supervised Contrastive Ordinal Loss for Abdominal Aortic Calcification Scoring on Vertebral Fracture Assessment Scans}

\titlerunning{Supervised Contrastive Ordinal Loss for AAC Scoring}


\author{Afsah Saleem\inst{1,2}\and Zaid Ilyas\inst{1,2} \and David  Suter\inst{1,2}\and Ghulam Mubashar Hassan \inst{3}  \and Siobhan Reid \inst{4} \and John T. Schousboe \inst{5} \and Richard Prince \inst{3} \and William D. Leslie \inst{6} \and Joshua R. Lewis \inst{1,2} \and Syed Zulqarnain Gilani \inst{1,2,3}}

\authorrunning{A. Saleem et al.}

\institute{Centre for AI \& ML, School of Science, Edith Cowan University,
Joondalup, Australia \and Nutrition and Health Innovation Research Institute, Edith Cowan University,
Joondalup, Australia \and 
Computer Science and Software Engineering, The University of Western Australia, Perth, Australia  \and 
Department of Electrical and Computer Engineering, University of Manitoba, Winnipeg, Canada \and 
Park Nicollet Clinic and HealthPartners Institute, HealthPartners, Minneapolis,USA \and 
Departments of Medicine and Radiology, University of Manitoba,
Winnipeg, Canada\\
\email{afsah.saleem@ecu.edu.au}}

\maketitle              

\begin{abstract}

Abdominal Aortic Calcification (AAC) is a known marker of asymptomatic Atherosclerotic Cardiovascular Diseases (ASCVDs). AAC can be observed on Vertebral Fracture Assessment (VFA) scans acquired using Dual-Energy X-ray Absorptiometry (DXA) machines.  Thus, the automatic quantification of AAC on VFA DXA scans may be used to screen for CVD risks, allowing early interventions.
In this research, we formulate the quantification of AAC  as an ordinal regression problem. We propose a novel Supervised Contrastive Ordinal Loss (SCOL) by incorporating a label-dependent distance metric with existing supervised contrastive loss to leverage the ordinal information inherent in discrete AAC regression labels.
We develop a Dual-encoder Contrastive Ordinal Learning (DCOL)  framework that learns the contrastive ordinal representation at global and local levels to improve the feature separability and class diversity in latent space among the AAC-24 genera. We evaluate the performance of the proposed framework using two clinical VFA DXA scan datasets and compare our work with state-of-the-art methods. Furthermore, for predicted AAC scores, we provide a clinical analysis to predict the future risk of a Major Acute Cardiovascular Event (MACE). 
Our results demonstrate that this learning enhances inter-class separability and strengthens intra-class consistency, which results in predicting the high-risk AAC classes with high sensitivity and high accuracy.

\end{abstract}

\section{Introduction} 
Abdominal Aortic Calcification (AAC) is an established marker of atherosclerotic cardiovascular disease (CVD) \cite{schousboe2017abdominal} and can help in identifying asymptomatic cases at risk for CVD-related hospitalizations and deaths \cite{lewis2018long}. CVDs are responsible for 32\% of global deaths, with atherosclerotic CVD events being the leading cause \cite{WHO}. Therefore, early detection and management of AAC can improve CVD prevention, and management \cite{lewis2018long}.
AAC can be seen on lateral spine Vertebral Fracture Assessment (VFA) images acquired using  Dual-energy X-ray Absorptiometry (DXA), X-rays or Computed Tomography (CT) \cite{leow2021prognostic,schousboe2017abdominal}. However, DXA imaging is the most recommended and cost-effective approach for VF assessment, with the least radiation exposure \cite{lems2021vertebral,yang2020vertebral}. In VFA DXA scans, AAC can be calculated manually using the Kauppila AAC-24 semi-quantitative scale \cite{kauppila1997new}. 
 However, the manual scoring of AAC in VFA images is arduous and subjective \cite{reid2021machine,schousboe2017abdominal}. 
Thus, designing an automated system for detecting and quantifying AAC in VFA DXA scans may be the most feasible approach for obtaining valuable CVD risk information in asymptomatic individuals.

Very few attempts have been made for automated AAC scoring in VFA DXA scans \cite{chaplin2019automated,elmasri2016automatic,gilani2022show,reid2021machine}.  
Reid et al. \cite{reid2021machine} trained two CNN models using VFA DXA scans and reported results via ensembling on a limited test set. 
Gilani et al. \cite{gilani2022show} performed sequential scoring using a vision-to-language model, but achieved low sensitivity for a moderate-risk class, indicating difficulty in handling complex cases near class boundaries. 
These works~\cite{gilani2022show,reid2021machine} considered automated AAC scoring a regression task. However, simple regression losses depend on continuous regression labels and, thus, cannot make feature embeddings separable. Moreover, inter-class similarities, intra-class variations and image artifacts in low-resolution VFA DXA scan further complicate the task. Therefore, using these losses directly for AAC-24 score regression may not be optimal.

Contrastive representation learning has shown promising results in medical image classification \cite{bhattacharya2022supervised,jaiswal2021scalp,li2021dual} and segmentation tasks \cite{chaitanya2020contrastive,hua2022uncertainty,zhao2022cross}. In the classification tasks, supervised contrastive learning (SupCon) \cite{khosla2020supervised} aims to bring feature embeddings with the same labels closer together in the latent space and move the dissimilar ones apart. However, SupCon cannot preserve the ordinal information of the regression labels in the latent space \cite{dai2021adaptive}. To address this, Dai et al. \cite{dai2021adaptive} propose supervised Adaptive Contrastive loss (AdaCon), which depends on an adaptive margin. For calculating the adaptive margin, they assumed that a regression label could be replaced with its Empirical Cumulative Distribution Function (ECDF) \cite{van2000asymptotic}. Though this assumption might be valid for large datasets, it may not work with highly skewed, imbalanced and limited-size datasets \cite{dai2021adaptive}.

 To this end, we propose a novel Supervised Contrastive Ordinal Loss (SCOL), considering AAC scoring as an ordinal regression problem. We integrate a label-dependent distance metric with the supervised contrastive loss.
Unlike AdaCon \cite{dai2021adaptive}, this metric relies exclusively on discrete regression labels, making it possible to utilize the ordinal information inherent in these labels. Using SCOL, we design an effective Dual-encoder Contrastive Ordinal Learning (DCOL) framework. Unlike previous methods, which either use global \cite{reid2021machine} or local attention-based features \cite{gilani2022show}, DCOL assimilates global and local feature embeddings to increase feature diversity, and class separability in latent space. 

To the best of our knowledge, this is the first framework that explores contrastive learning for automated detection of AAC.
Our contributions are summarized as follows: 1) We propose a novel supervised contrastive ordinal loss by incorporating distance metric learning with the supervised contrastive loss to improve inter-class separability and handle intra-class diversity among the AAC genera. 2) We design a Dual-Encoder Contrastive Ordinal Learning framework using the proposed loss to learn separable feature embeddings at global and local levels. 3) We achieve state-of-the-art results on two clinical datasets acquired using DXA machines from multiple manufacturers, demonstrating the generalizability and efficacy of our approach. 4) We compare the Major Adverse Cardiovascular Event (MACE) outcomes for machine-predicted AAC scores and the human-measured scores to explore the clinical relevance of our method.  
This work aims to contribute clinically in refining automated AAC prediction methods using low-energy VFA scans. Our code is available at \cite{AfsahS}.

\section{Methodology}

\begin{figure}[t]
\includegraphics[width=0.9\textwidth]{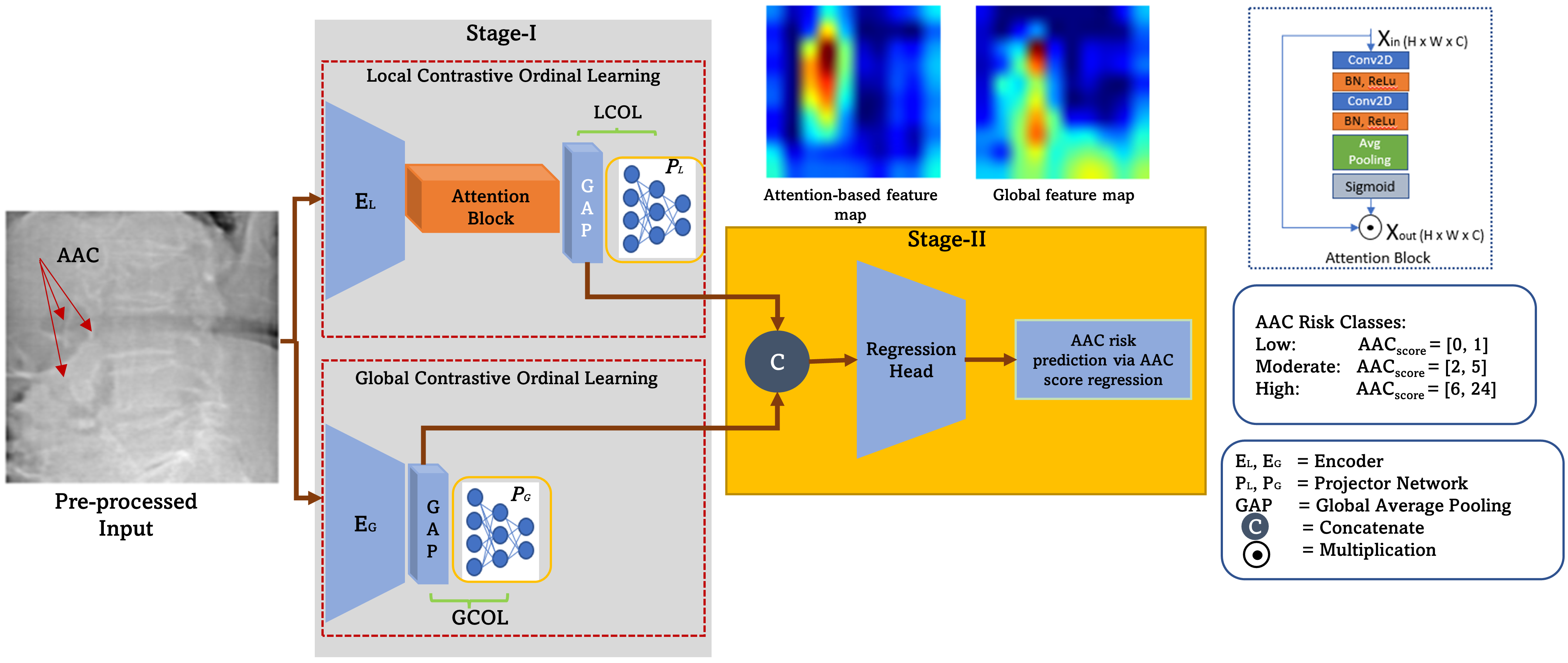}

\caption{Framework of our dual-encoder contrastive ordinal learning framework.} 
\label{fig1:framework}
\end{figure}

\subsection{Supervised Contrastive Ordinal Learning}

Consider a training set $T$ of $M$ image-label pairs, such that $T=\{(x_{i},y_{i})\}_{i=1}^{M}$ where $x_{i}$ is the $i_{th}$ VFA DXA scan and $y_{i}$ is the corresponding AAC score. Let $a$, $p$ and $n$ denote the indices of anchor, positive sample, and negative samples in a batch $I$.
$P(i)$ is the set of indices of all the positive samples, i.e., having the same AAC score as $y_{a}$ and $N(i)$ is the set of all other negative indices. 
Consider an encoder-projector network that maps the anchor image $x_{a}$ in the embedding space such that $z_{a}=Proj(Enc(x_{a}))$, then the similarity between any two projections $z_{i}$ and $z_{k}$ in the latent space is: $sim(z_{i},z_{k}) = z_{i}^{T}.z_{k}$.

Supervised contrastive loss \cite{khosla2020supervised}  pulls images of the same class (positive samples) close together and pushes the negative ones apart.
Following this strategy, we propose contrastive ordinal loss to move the negative sample $x_{n}$ apart from the anchor $x_{a}$ by a distance $r_{(a, n)}$: such that if $a<b<c$ then $r(a,b)$ must be less than $r(a,c)$ and vice versa. By incorporating this ordinal distance metric with the similarity function, we can maximize the benefit of ordinal information present in regression labels. Furthermore, in our AAC scoring task, this ordinal distance metric can help to increase inter-class separability and minimize the effect of intra-class variations.(For visualization see \textattachfile{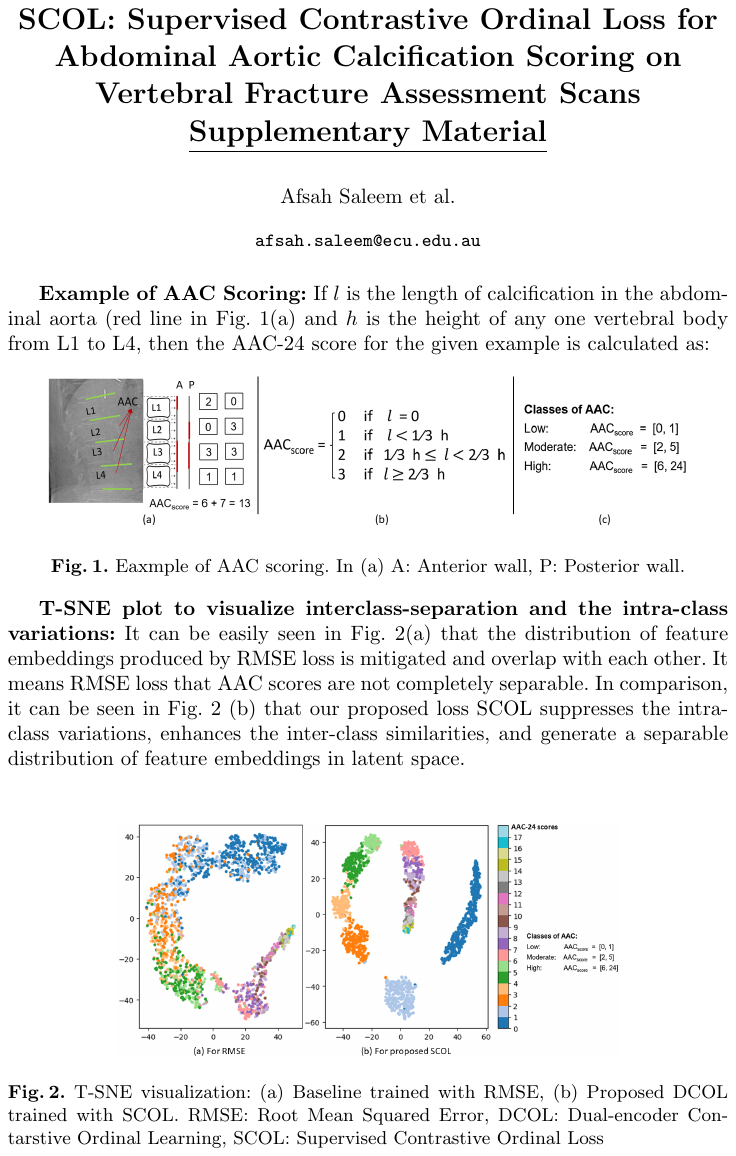}{\textcolor{black}{SM).}}  Inspired by \cite{dai2021adaptive}, we propose Supervised Contrastive Ordinal Loss (SCOL) for ordinal regression as:
\begin{equation}
L_{SCOL}=\sum_{i \in I}\frac{-1}{|P(i)|}\sum_{p\in P(i)}log\frac{exp(z_{a}.z_{p}/\tau)}{\sum_{n\in N(i)(n\neq a)} exp((z_{a}.z_{n} + r_{a,n}) /\tau)}
\label{eq:1}
\end{equation}
where $\tau$ is a scaling hyper-parameter for contrastive loss and $r_{(a, n)}$ is the distance metric between two labels $y_{(a)}$ and $y_{(n)}$. If C is the number of ordinal labels in training set T, then $r_{(a, n)}$ is calculated as:
$r_{a,n}=  ||y_{a} - y_{n}||_{2} \times 2/C$.

From the above equation, it can be seen that our ordinal distance metric is monotonically increasing, i.e., if $a<b<c$, then $r_{a,b}$ must be less than $r_{a,c}$ and vice versa. This property allows the metric to maintain the ordinality of the data while improving the class separability in the latent space.

\subsection{Dual-Enccoder Contrastive Ordinal Learning}
The proposed Dual-Encoder Contrastive Ordinal Learning (DCOL) framework consists of two stages: Stage-I: contrastive ordinal feature learning and Stage-II: relevant AAC risk class prediction via AAC-24 score regression (Fig. \ref{fig1:framework}). 

\textbf{Stage-I:} It consists of two modules: Local Contrastive Ordinal Learning (LCOL) and Global Contrastive Ordinal Learning (GCOL). In these modules, we train the global and local encoder-projector networks individually in an end-to-end manner to extract contrastive feature embeddings.

\textbf{\textit{Local Contrastive Ordinal Learning:}} In practice, to quantify AAC, clinicians focus on the aortic region adjacent to lumbar vertebrae L1-L4. Following this, in the Localized feature-based Contrastive Ordinal Learning module, $LCOL$, we integrate a simple yet effective localized attention block with the baseline encoder $E_{l}$ to roughly localize the aorta's position using only regression labels. This attention block is attached with $E_{l}$ after extracting the deep feature map $f_{m}$ from the last convolution layer. Our localized attention block consists of two 2D convolutional layers, followed by batch normalization and ReLu activation layers. This set of layers is then followed by an average pooling layer and sigmoid activation to create an activation map $f_{s}$ for the most salient features in the given image. Multiplying this activation map $f_{s}$ with the initial feature map $f_{m}$ results in extracting the most significant features from $f_m$. These features are then projected into the latent space for processing by our SCOL loss.
SCOL encourages the local contrastive embeddings with the same AAC score to move closer and the dissimilar ones apart based on the distance between their labels.

\textit{\textbf{Global Contrastive Ordinal Learning:}}
In the Global Contrastive Ordinal Learning module, we extract the global representation of a given VFA DXA scan. In encoder $E_{g}$, we replace the fully connected layers of the vanilla CNN model with a global average pooling $(GAP)$ layer for feature extraction. These feature embeddings are then passed to the projection network $P_{g}$. SCOL operates on projected embeddings extracted from the whole lumbar region to maximize the feature separability while preserving the ordinal information in latent space.
Both projector networks, $P_{l}$ and $P_{g}$, consist of two Dense layers with 1280 and 128 neurons each, followed by ReLu activation. 

\textbf{Stage-II:} In AAC-24 score regression, a small change in pixel-level information can move the patient from low to moderate or moderate to high-risk AAC class. Thus, to decrease the effect of intra-class variations and to increase the inter-class separability, we assimilate the features extracted from encoders $E_{l}$ and $E_{g}$. The resultant feature vector is fed as input to a feed-forward network consisting of two Dense layers with 1280 and 128 neurons each, followed by ReLu activation. Finally, a linear layer predicts the final AAC regression score. This module is trained using root mean squared error loss $L_{rmse}$ calculated as: $L_{rmse}=  \sqrt{\sum_{i \in m} ({y'_{i}-y_{i}})^{2}/m}$, where $m$ is the number of samples, $y_{i}$ are actual and $y'_{i}$ are predicted AAC scores. 
The resulting AAC scores are then further classified into three AAC risk classes. 

\section{Experiments and Results}
\subsubsection{Dataset and Annotations:}
\label{AAC_scoring}
We conducted experiments on two de-identified clinical datasets acquired using the Hologic 4500A and GE iDXA scanners. Both datasets are manually annotated by clinicians using the AAC-24 scale \cite{kauppila1997new} which divides the aortic walls into four segments based on the lumbar vertebrae bodies $(L1-L4)$.
Each segment is assigned an AAC score 0: if no calcification (Cal), 1: Cal $\leq 1/3$ of aortic wall, 2: Cal $> 1/3$ but $< 2/3$ of aortic wall and 3: Cal $ \geq 2/3$ of aortic wall.
The total AAC-24 score can range from $0$ to $24$ and is further classified into three AAC risk classes using clinical thresholds \cite{lewis2016abdominal}: Low-risk $(AAC < 2)$, Moderate-risk $(2 \leq AAC \leq 5)$, and High-risk $(AAC > 5)$.

The \textbf{Hologic Dataset} \cite{lewis2018long} has 1,914 single-energy DXA scans acquired using a Hologic 4500A machine. Each scan has dimensions of at least $800 \times 287$ pixels. Among 1,914 scans, there are $764$ scans belonging to the low-risk, $714$ to moderate-risk and $436$ scans to the high-risk AAC group. The \textbf{iDXA GE Dataset} \cite{reid2021machine,gilani2022show} has 1,916 dual-energy VFA DXA scans. Among these, there are $829$ belonging to low-risk, $445$ to moderate-risk and $642$ scans to the high-risk AAC group. These scans are acquired using an iDXA GE machine. These scans have dimensions of $1600 \times 300$ pixels.

\textbf{Implementation Details:} Each VFA scan in both datasets contains a full view of the thoracolumbar spine. To extract the region of interest (ROI), i.e., the abdominal aorta near the lumbar spine, we crop the upper half of the image, resize it to $300 \times 300$ pixels and rescale it between $0$ and $1$.
We apply data augmentations including rotation,  shear and translation. 
We implement all experiments in TensorFlow \cite{Tensorflow}, using stratified 10-fold cross-validation on a workstation with NVIDIA RTX 3080 GPU. (For details, see \textattachfile{Paper1923Supplementary.pdf}{\textcolor{black}{SM.)}}
In stage-I, we adopted an efficient, smaller and faster pre-trained model, EfficientNet-V2S \cite{tan2021efficientnetv2}, as the backbone for both encoders. 
We train stage-I for $200$ epochs and stage II for $75$ epochs using the RMSprops optimizer with the initial learning rate of $3 \times 10^{-4}$ and a batch size of $16$. $\tau$ in the proposed $SCOL$ is $0.2$. The inference time for each scan is less than $15$ ms. 
To avoid overfitting, we used early stopping and reduce LR on the plateau while optimizing the loss.


\textbf{Evaluation Metrics:} The performance of the AAC regression score is evaluated in terms of Pearson's correlation, while for AAC risk classification task Accuracy, F1-Score, Sensitivity, Specificity, Negative Predictive Value (NPV) and Positive Predictive Value (PPV) are used in One Vs. Rest (OvR) setting. 
\textbf{Baseline:} 
EfficientNet-V2S model trained in regression mode using RMSE loss.

\begin{table}[tb]
\caption{Comparison of proposed loss SCOL with different losses using Hologic Dataset}\label{ablation}

\centering
\begin{tabular}{|p{3 cm}|p{0.8cm}|p{0.8cm}|p{1.2cm}|p{1.3cm}|p{1.3cm}|p{1.45cm}|p{1.45cm}|}
\hline
Loss& \multicolumn{2}{|c|}{Method} &Pearson&  Accuracy & F1-Score & Sensitivity &Specificity\\ \cline{2-3}
& GCL & LCL &(\%)& (\%)& (\%)& (\%) &(\%) \\ \hline
RMSE  &\checkmark & -                 & 87.02 & 81.92 & 75.88 & 74.93 & 85.36\\ 
RMSE  &- & \checkmark &87.34 & 82.5& 76.71 & 76.06 & 85.88 \\ 
RMSE &\checkmark & \checkmark &88.12 & 83.52&  77.91 &77.00&86.64\\ \hline
SupCon\cite{khosla2020supervised}&\checkmark &-& 87.26 & 83.52 & 77.97 & 77.14 & 86.64 \\
SupCon\cite{khosla2020supervised} & -& \checkmark  &86.39& 83.38 & 77.74 & 77.02 & 86.60 \\
SupCon \cite{khosla2020supervised}&  \checkmark & \checkmark  & 88.24& 83.94&78.69 &77.82 &87.00\\ \hline
AdaCon \cite{dai2021adaptive}&  \checkmark &- & 88.14&82.51& 76.67  &75.82 &85.86\\
AdaCon \cite{dai2021adaptive}& - & \checkmark  &88.60&  83.69& 78.16  &77.28 &86.79\\ 
AdaCon \cite{dai2021adaptive}&  \checkmark & \checkmark  & 88.40 &  83.98& 78.80  &77.91 &87.03\\ \hline
SCOL (This paper)& \checkmark&  - & 88.74& 84.08 & 78.82 & 78.13 & 87.10 \\ 
SCOL (This paper) & - & \checkmark  &88.92&  84.36 & 78.90 & 77.93 & 87.34 \\ 
SCOL (This paper)&  \checkmark  & \checkmark& \textbf{89.04}& \textbf{85.27 }& \textbf{80.25} & \textbf{79.46} & \textbf{88.05}\\ 
\hline
\end{tabular}
\end{table}

\begin{table}[tb]
\caption{Class-wise performance comparison of the proposed framework with baseline. NPV: Negative Predicted Value, PPV: Positive Predicted Value}

\centering
\begin{tabular}{|p{0.5cm}|p{1.4cm}|p{1.2cm}|p{1.3cm}|p{1.3cm}|p{1cm}|p{1cm}|p{1.5cm}|p{1.5cm}|}
\hline
&AAC  & Method&  Accuracy & F1-Score & NPV & PPV & Sensitivity &Specificity\\ 

&Class & &  (\%) & (\%) & (\%) & (\%) & (\%) & (\%) \\
\hline

\multirow{8}{*}{\begin{turn}{90}\textbf{Hologic Dataset}\end{turn}}&Low &Baseline&  75.96 & 69.00  &78.89 & 71.11& 67.01 &81.91\\
&(n=764)&DCOL  &\textbf{80.56 }& \textbf{75.52 }& \textbf{83.59} & \textbf{75.93} & \textbf{75.13 }&\textbf{84.17}\\ 
\cline{2-9}
&Moderate&Baseline&72.94   & 65.92 & 80.78& 62.16  & 70.17 & 74.58 \\ 
&(n=714)&DCOL  & \textbf{77.90} & \textbf{71.36} & \textbf{83.75} & \textbf{69.06} & \textbf{73.80} & \textbf{80.33} \\ 
\cline{2-9}
&High &Baseline  &96.89 & 92.71  & 96.46 & 98.45& 87.61 & 99.50 \\ 
&(n=436)&DCOL &\textbf{97.33} &\textbf{93.86} & \textbf{96.97} & \textbf{98.73} & \textbf{89.45} &\textbf{99.66}\\ 
\cline{2-9}
&Average &Baseline& 81.92 & 75.88 & 77.24 & 85.38 & 74.93 & 85.36\\ 
&(n=1,914)&DCOL  &\textbf{85.27} & \textbf{80.25} &\textbf{81.24} & \textbf{88.10 }& \textbf{79.46} & \textbf{88.05}\\  \hline
\multirow{8}{*}{\begin{turn}{90}\textbf{iDXA GE Dataset}\end{turn}}&Low &Baseline  &85.44 & 82.97 & 83.95 & 86.53 & 82.02 & 88.04\\ 
&(n=829)&DCOL  &\textbf{87.63} & \textbf{85.76 }& \textbf{85.40} & \textbf{89.35} & \textbf{86.12 }& 88.37\\ 
\cline{2-9}
&Moderate & Baseline& 77.60 & 55.35 & 51.55 & 87.21 & 59.77 & 83.00\\ 
&(n=445)&DCOL & \textbf{80.63} & \textbf{60.15} & \textbf{57.61 }& \textbf{88.46} & \textbf{62.92} &\textbf{85.99} \\ 
\cline{2-9}
&High &Baseline & 90.18 & 84.74 & 88.47 & 90.95 & 81.30 & 94.66 \\ 
&(n=642)&DCOL & \textbf{90.92} & \textbf{85.99} & \textbf{89.39} & \textbf{91.60} & \textbf{82.71 }& \textbf{95.05}\\ 
\cline{2-9}
&Average &Baseline &84.41 & 74.35 & 74.65 & 88.23 & 74.37 & 88.56 \\ 
&(n=1,916)&DCOL   & \textbf{86.39} & \textbf{77.28} & \textbf{77.47} & \textbf{89.80} & \textbf{77.25} & \textbf{89.94}\\\hline
\end{tabular}

\label{res:CAIFOS}
\end{table}

\begin{table}[tb]
\caption{Comparison of the proposed DCOL with State-of-the-Art methods \cite{reid2021machine,gilani2022show} on iDXA GE Dataset. NPV: Negative Predicted Value, PPV: Positive Predicted Value}

\centering
\begin{tabular}{|p{1.5cm}|p{3 cm}|p{1.1cm}|p{1.3cm}|p{1cm}|p{1cm}|p{1cm}|p{1cm}|}
\hline
Risk Class&Method & Pearson&Accuracy  &  Sens. & Spec. & NPV  & PPV \\ 
& &  (\%)& (\%)  &  (\%) & (\%)& (\%)  & (\%) \\ 
\hline
Low & Reid et al. \cite{reid2021machine} &-&71.14 & 55.49 & 83.07 & 70.99 & 71.43 \\ 
&Gilani et al. \cite{gilani2022show}  &-&82.52 & \textbf{86.37} & 79.58 & 88.45 & 76.33  \\ 
& DCOL (This paper)   &-&  \textbf{87.63} & 86.12 & \textbf{88.77} & \textbf{89.35} &\textbf{85.40}\\
\hline
Moderate & Reid et al. \cite{reid2021machine}&-& 62.06 & 59.33 & 62.88 & 83.63 & 32.59  \\ 
&Gilani et al. \cite{gilani2022show}  &-& 75.52 & 37.53 & \textbf{87.02} & 82.16 & 46.65  \\
&DCOL (This paper)   &-&  \textbf{80.63} &\textbf{62.90 }& 85.99 & \textbf{88.46} & \textbf{57.60} \\ 
\hline 
High &Reid et al. \cite{reid2021machine} &-& 79.12 & 54.83 & 91.37 & 80.06 & 76.19  \\ 
&Gilani et al. \cite{gilani2022show}  &-&87.89 & 80.22 & 91.76 & 90.20 & 83.06   \\ 
&DCOL (This paper)   &-& \textbf{90.90 }& \textbf{82.70} & \textbf{95.05} & \textbf{91.60 }& \textbf{89.39}\\ 
\hline
Average &Reid et al. \cite{reid2021machine} & 65.00 &  70.77 & 56.55 & 79.11 & 78.23 & 60.07\\ 
&Gilani et al. \cite{gilani2022show}   &84.00 & 81.98 & 68.04 & 86.12 & 86.93 & 68.68  \\ 
&DCOL (This paper)      & \textbf{91.03}&\textbf{86.39} & \textbf{77.25} & \textbf{89.94 }& \textbf{89.80} &\textbf{77.47 }\\ 
\hline
\end{tabular}

\label{res:comparison}
\end{table}
\textbf{Ablation Study:}
Table \ref{ablation} highlights the efficacy of our proposed loss $SCOL$. We train our dual-encoder contrastive learning framework with proposed SCOL, SupCon \cite{khosla2020supervised} and AdaCon \cite{dai2021adaptive}, individually, on the Hologic dataset. We also evaluate the performance of the local and global contrastive modules (LCL and GCL) with each contrastive loss. Table \ref{ablation} also shows the strength of integrating the localized attention block with the baseline model trained 
with RMSE loss.

\textbf{Comparison with the Baseline:} 
 In Table \ref{res:CAIFOS}, we compare the performance of our framework with the baseline on both datasets. 
For the Hologic dataset, our proposed method improved Pearson's correlation coefficient from a baseline of 0.87 to 0.89 and 3-class classification accuracy from $73\% \pm 3.82$ to $78\% \pm 3.65$ with $(p < 0.001)$. While, for the iDXA GE dataset, the proposed method enhanced the Pearson's correlation from a baseline of 0.89 to 0.91 and averaged 3-class classification accuracy from $77\% \pm 3.9 $ to $80\% \pm 5.12$ with $(p < 0.001)$. 
\begin{figure}[tb]
     \centering
     \includegraphics[width=0.75
     \textwidth]{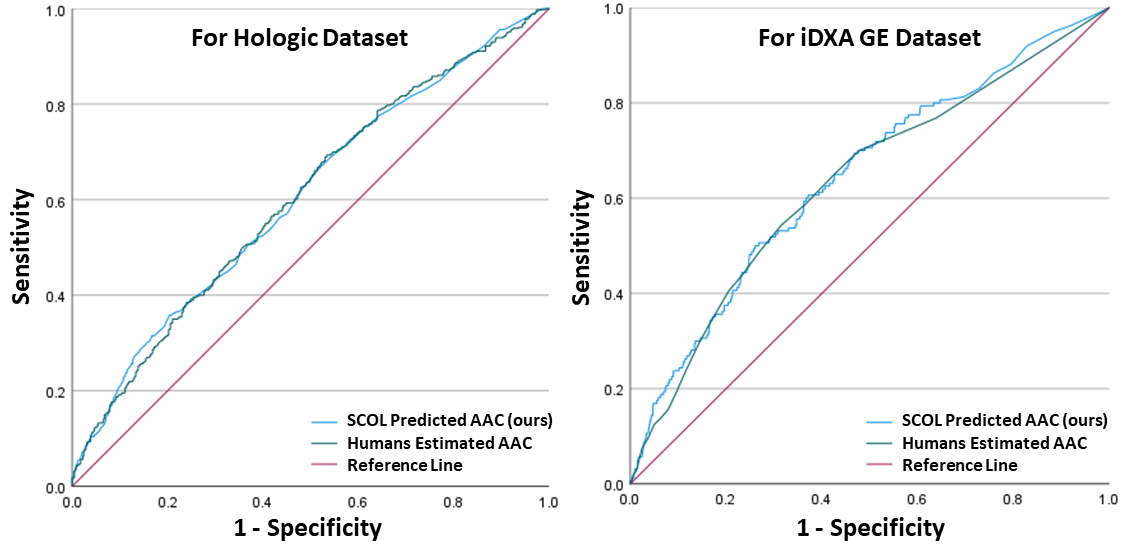}
\caption {Comparison of AUCs for Major Adverse Cardiovascular Events (MACE) associated with our predicted AAC-24 scores and human-measured AAC-24 scores.}
\label{ROC}
\end{figure}

\textbf{Comparison with the State-of-the-Art (SOTA):} 
Table~\ref{res:comparison} shows the comparison of our proposed framework with two SOTA methods \cite{reid2021machine,gilani2022show} using the iDXA GE dataset.
Our approach outperforms \cite{reid2021machine} by an average of 15.62\% in accuracy and 20.7\% in sensitivity with $(p < 0.001)$, while in comparison to \cite{gilani2022show}, accuracy is improved by 4.41\% and sensitivity by 9.21\%, with $(p < 0.001)$. %

\textbf{Clinical Analysis and Discussion:} 
To signify the clinical significance, we estimate the AUCs (Fig. \ref{ROC}) for Major Adverse Cardiovascular Events (MACE) associated with our predicted AAC-24 scores versus human-measured AAC-24 scores.
For the Hologic dataset, in the cohort of 1083 women with clinical follow-up, 460 suffered a MACE event. Predicted AAC-24 scores provided similar AUCs to human AAC-24 scores (0.61 95\%CI 0.57-0.64 vs. 0.61 95\%CI 0.57-0.64).
The three predicted AAC groups had 598 (55.2\%), 394 (30.0\%) and 159 (14.7\%)  of women in the low, moderate, and high AAC groups, respectively, with MACE events occurring in 37.8\%, 46.2\%, and 52.8\% of these groups, respectively. 
The age-adjusted HRs for MACE events
in the moderate and high AAC groups were 1.24 95\%CI 1.01-1.53 and 1.45 95\% CI 1.13-1.87, respectively, compared to the low predicted AAC group. These HRs were similar to the human AAC groups, i.e., for moderate
and high AAC groups, HR 1.28 95\%CI 1.04-1.57 and 1.47 95\% CI 1.15-1.89, respectively, compared to the human low AAC group. 

For the iDXA GE Dataset, in the cohort of 1877 patients with clinical follow-up, 160  experienced a MACE event. The AUCs of predicted AAC-24 scores were similar AUC to human AAC-24 (0.64 95\%CI 0.60-0.69 vs. 0.63 95\%CI 0.59-0.68). 
The predicted AAC groups had 877 (46.7\%), 468 (24.9\%), and 532 (28.3\%) of people in the low, moderate, and high AAC groups, respectively, with MACE events occurring in 5.1\%, 7.5\%, and 15.0\%  of these groups, respectively. 
The age and sex-adjusted HR for MACE in the moderate AAC group was 1.21 95\%CI 0.77-1.89, and 2.64 95\% CI 1.80-3.86 for the high AAC group, compared to the low predicted AAC group, which were similar to the HRs of human AAC groups, i.e., for moderate and high AAC groups HR 1.15 95\%CI 0.72-1.84 and 2.32 95\% CI 1.59-3.38, respectively, compared to the human low AAC group.
\section{Conclusion}
We propose a novel Supervised Contrastive Ordinal Loss and developed a Dual-encoder Contrastive Ordinal Learning framework for AAC scoring and relevant AAC risk classification in low-energy VFA DXA scans. Our framework learns contrastive feature embeddings at the local and global levels. 
Our results demonstrate that the contrastive ordinal learning technique remarkably enhanced inter-class separability and strengthened intra-class consistency among the AAC-24 genera, which is particularly beneficial in handling challenging cases near the class boundaries.
Our framework with SCOL loss demonstrates significant performance improvements, compared to state-of-the-art methods. 
Moreover, the ablation studies also establish the effectiveness of our dual-encoder strategy and localized attention block.  
These results suggest that our approach has great clinical potential for accurately predicting AAC scores and relevant risk classes.
\subsubsection*{Acknowledgement and Data Use Declaration:}
De-identified labelled images were sourced for the ML (Project no: 03349 LEWIS) from a number of existing studies collecting VFAs.  For Hologic dataset, written informed consent was obtained from all participants. 
The Human Ethics Committee of the University of Western Australia approved the study protocol and consent form (approval no. 05/06/004/H50). The Human Research Ethics Committee of the Western Australian Department of Health also approved the data linkage study (approval no. 2009/24). For the GE images the study was approved by the Health Research Ethics Board for the University of Manitoba (HREB H2004:017L, HS20121). The Manitoba Health Information Privacy Committee approved access to the iDXA GE data and waived the requirement for signed consent (HIPC 2016/2017–29). 


The study was supported by a National Health and Medical Research Council of Australia Ideas grant (APP1183570) and the Rady Innovation Fund, Rady Faculty of Health Sciences, University of Manitoba.
The results and conclusions are those of the authors and no official endorsement by Manitoba Health and Seniors Care, or other data providers is intended or should be
inferred.
The salary of JRL is supported by a National Heart Foundation of Australia Future Leader Fellowship (ID: 102817). Also, SZG was partially funded by the Raine Priming Grant awarded by Raine Medical Research Foundation.

%

\bibliographystyle{splncs04}
\bibliography{Paper1923}

\end{document}